\pdfoutput=1
\documentclass[10pt,twocolumn,letterpaper]{article}

\usepackage[dvipsnames]{xcolor}
\usepackage{iccv}
\usepackage{times}
\usepackage{epsfig}
\usepackage{graphicx}
\usepackage{amsmath}
\usepackage{amssymb}
\usepackage{siunitx}
\usepackage{bbold}
\usepackage{lipsum}
\usepackage{ textcomp }

\usepackage{stfloats}
\usepackage{multirow}
\usepackage{multicol}
\usepackage{tabularx}
\usepackage{caption} 
\usepackage{subcaption}
\usepackage[normalem]{ulem}
\useunder{\uline}{\ul}{}

\usepackage[nodisplayskipstretch]{setspace}
\usepackage[pagebackref=true,breaklinks=true,letterpaper=true,colorlinks,bookmarks=false]{hyperref}

\iccvfinalcopy 

\def\iccvPaperID{4123} 

\begin{document}

\title{Monocular 3D Object Detection via Geometric Reasoning on Keypoints}

\author{Ivan Barabanau \\
Skoltech\\
{\tt\small ivan.barabanau@skoltech.ru}
\and
Alexey Artemov \\
Skoltech\\
{\tt\small a.artemov@skoltech.ru}
\and
Evgeny Burnaev \\
Skoltech\\
{\tt\small e.burnaev@skoltech.ru}
\and
Vyacheslav Murashkin \\
Yandex\\
{\tt\small vmurashkin@yandex-team.ru}
}

\maketitle


\newcommand{\EB}[1]{}
\newcommand{\IB}[1]{}
\newcommand{\LA}[1]{}
\newcommand{\todo}[1]{}

\newcommand{\ra}[1]{\renewcommand{\arraystretch}{#1}}

\ra{1.04}
\setlength{\tabcolsep}{5.5pt}

\newcommand{\Argmax}{\mathop{{\rm Arg\,max}}}
\newcommand{\Argmin}{\mathop{{\rm Arg\,min}}}
\newcommand{\argmax}{\mathop{{\rm arg\,max}}}
\newcommand{\argsup}{\mathop{{\rm arg\,sup}}}
\newcommand{\argmin}{\mathop{{\rm arg\,min}}}

\newcommand{\lowresdepth}{d^{\mathrm{(LR)}}}
\newcommand{\hiresdepth}{d^{\mathrm{(HR)}}}
\newcommand{\hiresdepthest}{\widehat{d}^{\mathrm{(HR)}}}
\newcommand{\hiresimage}{I}
\newcommand{\hiresimageex}{I^{\mathrm{(HR)}}}

\def\vs.{vs.\spacefactor=\the\sfcode`\v}
\def\etc.{etc.\spacefactor=\the\sfcode`\c}

\begin{abstract}
Monocular 3D object detection is well-known to be a challenging vision task due to the loss of depth information; attempts to recover depth using separate image-only approaches lead to unstable and noisy depth estimates, harming 3D detections. 
In this paper, we propose a novel keypoint-based approach for 3D object detection and localization from a single RGB image. We build our multi-branch model around 2D keypoint detection in images and complement it with a conceptually simple geometric reasoning method. 
Our network performs in an end-to-end manner, simultaneously and interdependently estimating 2D characteristics, such as 2D bounding boxes, keypoints, and orientation, along with full 3D pose in the scene. We fuse the outputs of distinct branches, applying a reprojection consistency loss during training. 
The experimental evaluation on the challenging KITTI dataset benchmark demonstrates that our network achieves state-of-the-art results among other monocular 3D detectors.


\end{abstract}

\section{Introduction}

The success of autonomous robotics systems, such as self-driving cars, largely relies on their ability to operate in complex dynamic environments; as an essential requirement, autonomous systems must reliably identify and localize non-stationary and interacting objects, \eg vehicles, obstacles, or humans. In its simplest formulation, localization is understood as an ability to detect and frame objects of interest in 3D bounding boxes, providing their 3D locations in the surrounding space. Crucial to the decision-making process is the accuracy of depth estimates of the 3D detections.

Depth estimation could be approached from both hardware and algorithmic perspectives. On the sensors end, laser scanners such as LiDAR devices have been extensively used to acquire depth measurements sufficient for 3D detection in many cases~\cite{Yang_2018_CVPR, cvpr17chen, lang2018pointpillars, Zhou_2018_CVPR, ku2017joint, yan2018second}. However, point clouds produced by these expensive sensors are sparse, noisy and massively increase memory footprints with millions of 3D points acquired per second. 
In contrast, image-based 3D detection methods offer savings on CPU and memory consumption, use cheap onboard cameras, and work with a~wealth of established detection architectures (\eg,~\cite{liu2015ssd,redmon2015look,ren2015faster,lin2016feature,he2017mask,lin2017focal,liu2018path}), yet they require 
sophisticated algorithms for depth estimation, as raw depth cannot be accessed anymore. 


Recent research on monocular 3D object detection relies on separate dense depth estimation models~\cite{qin2018monogrnet, xu2018multi}, but depth recovery from monocular images is naturally ill-posed, leading to unstable and noisy estimates. In addidion, in many practical instances, \eg, with sufficient target resolution or visibility, dense depth estimation might be redundant in context of 3D detection. Instead, one may focus on obtaining sparse but salient features, such as 2D keypoints, that are well-known visual cues often serving as geometric constrains in various vision tasks such as human pose estimation~\cite{ramakrishna2012reconstructing,martinez2017simple,mehta2017monocular,VNect_SIGGRAPH2017} and more general object interpretation~\cite{hejrati2012analyzing,wu2016single}. 

Motivated by this observation, in this paper we propose a novel keypoint-based approach for 3D object detection and localization from a single RGB image. We build our model around 2D keypoint detection in images and complement it with a conceptually simple geometric reasoning framework, establishing correspondences between the detected 2D keypoints and their 3D counterparts defined on surfaces of 3D CAD models. The framework operates under the general assumptions, assuming the camera intrinsic parameters are given, and retrieves depth of closest keypoint instance, thereby "lifting" 2D keypoints to 3D space; the remaining 3D keypoints and the final 3D detection are assembled in a similar way. Our approach does not require keypoint-annotated labeled images, but instead relies on a multi-task reprojection consistency loss, allowing for robust 2D keypoint detection. Thus, our model is end-to-end trainable. 




In summary, our contributions are as follows:
\begin{itemize}
    \item We propose a novel deep learning-based framework for monocular 3D object detection, combining well-established region-based detectors and a geometric reasoning step over keypoints. 
    
    \item We describe an end-to-end training scheme for this framework, using a dataset of real-world images and a collection of 3D CAD models, annotated with 3D keypoints.
    
    
\end{itemize}


The rest of this paper is organized as follows. In Section~\ref{sec:related}, we review the related work on object detection, mostly in the context of self-driving and robotics applications. In Section~\ref{sec:framework}, we describe our proposed monocular 3D object detection approach, and in Section~\ref{sec:exper}, its experimental evaluation using the standard KITTI benchmark. We conclude in Section~\ref{sec:conclusion} with a discussion of our results.

\section{Related work}
\label{sec:related}


\paragraph{2D object detection. }

2D object detection is an extensively studied vision task, with a body of research devoted to both algorithms~\cite{redmon2015look,liu2015ssd,lin2017focal,ren2015faster,he2017mask,liu2018path} and  benchmarks~\cite{imagenet_cvpr09,Everingham10thepascal,lin2014microsoft,geiger2012we,geiger2013vision}. Traditionally, object detectors operate in two stages, with the first stage selecting object candidates~\cite{uijlings2013selective,zitnick2014edge,ren2015faster} and the second stage operating as a discriminator and refinement model, rejecting bad proposals~\cite{ren2015faster,he2017mask,dai2016r}. Due to the introduction of novel backbone architectures~\cite{xie2017aggregated} and losses~\cite{lin2017focal}, such approaches have attained top results in a number of benchmarks.
In the context of robotics, single-stage detectors such as YOLO~\cite{redmon2015look}, SSD~\cite{liu2015ssd} and RetinaNet~\cite{lin2017focal} are of particular interest, however, they offer inferior performance and are not straightforward to extend to related tasks such as instance segmentation~\cite{he2017mask, liu2018path} and keypoint detection~\cite{he2017mask}.

\paragraph{3D object detection. }

Recently, novel deep learning architectures operating directly on unstructured point clouds have been proposed~\cite{qi2017pointnet,qi2017pointnet++,dgcnn,pcpnet,pwcnn,pointcnn}, offering the possibility to develop corresponding 3D object detectors~\cite{qi2018frustum,Zhou_2018_CVPR,xu2018pointfusion}. However, such approaches require expensive sensing equipment (LiDARs) and commonly process point cloud data coupled with RGB data.
Some depth-based approaches operate over voxel-grid representations of the point clouds, leveraging the existing convolutional architectures~\cite{Yang_2018_CVPR,lang2018pointpillars,Zhou_2018_CVPR,yan2018second}, while other methods fuse depth features with birds-eye-view (BEV) and image features~\cite{cvpr17chen, ku2017joint, qi2018frustum}.

\paragraph{Monocular 3D object detection. }

The most relevant to our work is research on monocular 3D object detection, that is well-known to be a challenging vision task. 
Deep3DBox~\cite{MousavianCVPR2017} relies on a set of geometric constraints between 2D and predicted 3D bounding boxes and reduces 3D object localization problem to a linear system of equations, fitting 3D box projections into 2D detections. Their approach relies on a separate linear solver; in contrast, our model is end-to-end trainable and does not require external optimization.
Mono3D~\cite{Chen2016CVPR} extensively samples 14K 3D bounding box proposals per image and evaluates each, exploiting semantic and image-based features. In contrast, our approach does not rely on an exhaustive sampling in 3D space, bypassing a significant computational overhead.
OFT-Net~\cite{roddick2018orthographic} introduces an orthographic feature transform which maps RGB image features into a birds-eye-view representation through a 3D scene grid, solving the perspective projection problem. However, back-projecting image features onto 3D grid results in a coarse feature assignment. Our approach detects 2D keypoints with sufficient precision, avoiding any additional discretization.
MonoGRNet~\cite{qin2018monogrnet} directly deals with depth estimation from a single image, training an additional sub-network to predict the $z$-coordinate of each 3D bounding box. 
\cite{xu2018multi} exploit a similar approach, estimating disparity using a stand-alone pre-trained MonoDepth network~\cite{godard2017unsupervised}. Both these methods rely on the non-trainable depth estimation networks, which introduce a computational overhead; in contrast, our approach jointly estimates object 2D bounding-box and 3D pose in a fully trainable manner, not requiring a dense depth prediction.

Perhaps, the most similar approach to ours is~\cite{chabot2017deep}, which utilizes 3D CAD models, along with predicting 2D keypoints. However, their network only models 2D geometric properties and aims at matching the predictions to one of the CAD shapes, while 3D pose estimation is postponed for the inference step. They additionally exploit extensive annotations of keypoints in their 3D models. In contrast, we only annotate 14 keypoints per each of the five 3D models and exploit them in a geometric reasoning module to bridge the gap between 2D and 3D worlds, which allows us to deal with 3D characteristics during training in an end-to-end manner.


\begin{figure*}[t!]
 \centering
 \includegraphics[width=\textwidth, height=\textheight,keepaspectratio]{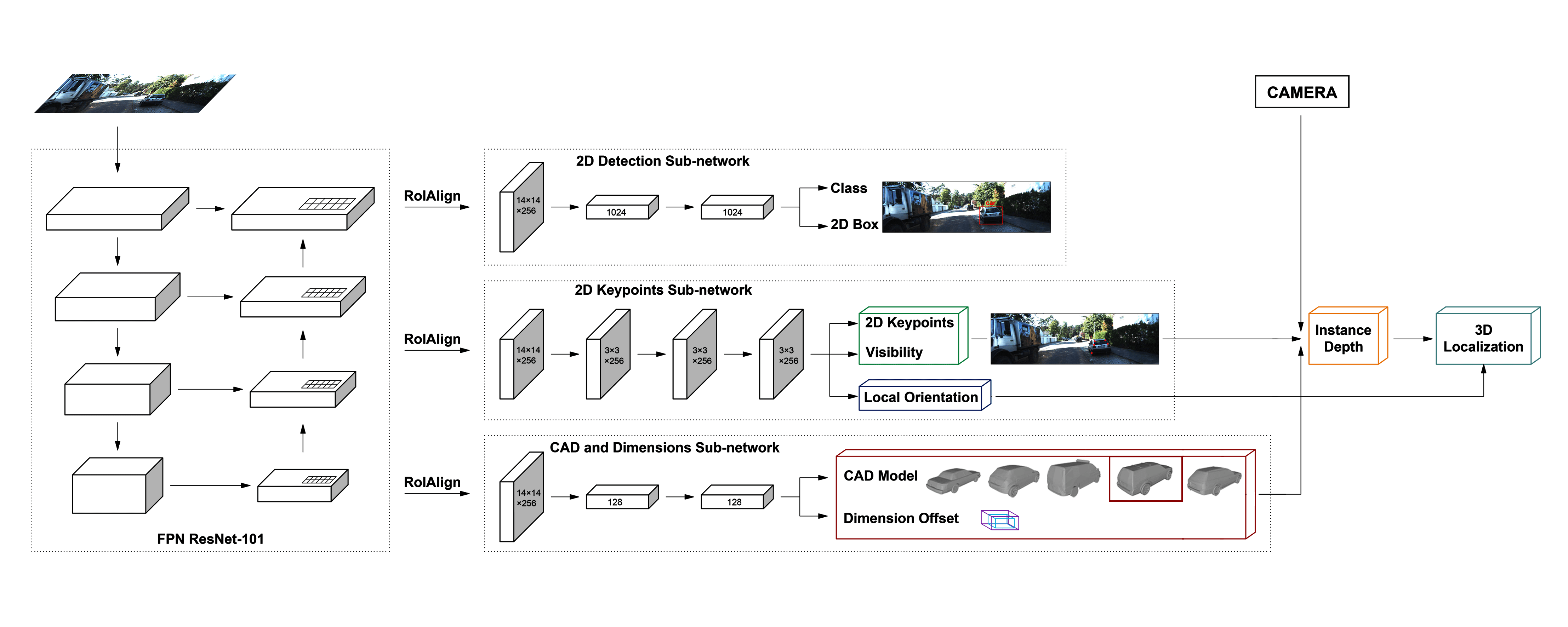}
 \caption{An overview of our monocular 3D detection architecture. We start with a universal backbone network (see use Mask R-CNN~\cite{he2017mask}) and complement it with three sub-networks: 2D object detection sub-network, 2D keypoints regression sub-network, and dimension regression sub-network. The network is trained end-to-end using a multi-task loss function.}
 \label{fig:pipeline}
\end{figure*}

\paragraph{Keypoints estimation and 3D representations. } 
Keypoint-based representations are a common mechanism of encoding 3D geometric structure in objects, that have proven themselves as powerful visual cues for tasks such as pose estimation~\cite{ramakrishna2012reconstructing,martinez2017simple,mehta2017monocular,VNect_SIGGRAPH2017}, fine pose prediction~\cite{lim2013parsing}, 3D reconstruction~\cite{snavely2006photo}, shape alignment~\cite{lim2013parsing,marion2018label}, to name a few. A commonly used approach is to learn a set of keypoint detectors, followed by some post-processing to assemble their predictions into a geometric model. However, obtaining sufficient ground-truth for training keypoint detectors is a challenging task. One may manually annotate 3D keypoints of objects in real images, but this is labor-intensive and often inaccurate. Other directions involve active shape modeling~\cite{ramakrishna2012reconstructing,lin2014jointly} and shape alignment with wireframe~\cite{zia2011revisiting,zia2013detailed,zeeshan2013explicit} and 3D CAD models~\cite{chabot2017deep}. 
For human pose estimation, another option could be motion capture of joint locations~\cite{martinez2017simple}
Recently, latent modeling approaches have been proposed to learn optimal sets of keypoints without direct supervision~\cite{wu2016single,suwajanakorn2018discovery}. Our keypoint detection approach bears similarity to~\cite{chabot2017deep} as we utilize 3D CAD models and align them to sensor measurements, but offers labor savings since we only annotate 14 keypoints per each of the five CAD models.


\section{3D Object Detection Framework}
\label{sec:framework}

Given a single RGB image, our goal is to localize target objects in the 3D scene. To do this, we propose an end-to-end trainable CNN-based framework that accepts a single RGB image as input and outputs a set of 3D detections. 
Each target object is defined by its class and 3D bounding box, parameterized by 3D center coordinates $\textbf{C} = (c_x, c_y, c_z)$ in a camera coordinate system, global orientation $\textbf{R} = (\theta, \cdot, \cdot)$, and dimensions $\textbf{D} = (w, h, l)$, standing for width, height and length, respectively (we don't correct for truncation or occlusion when defining object sizes). We parameterize global object orientation with the yaw angle $\theta$ only, which is a commonly adopted premise when dealing with objects in the road scenes~\cite{MousavianCVPR2017,qi2018frustum}. 


Our proposed framework comprises two sub-modules, each of which operates on the characteristics living in either 2D (image) or 3D (world) space. From the 2D perspective, each object of interest, cropped by its predicted 2D bounding box, is provided with 2D keypoints and their respective visibility states. On the 3D side, object dimensions, 3D CAD model, and local orientation are predicted. 
The gap between the two spaces is bridged by the geometric reasoning module computing instance depth, global orientation, and the final 3D detection.



Our implementation takes advantage of the generality of the state-of-the-art Mask R-CNN architecture~\cite{he2017mask}, viewing it as a universal backbone network extensible to adjacent problems, and complement it with three sub-networks: 2D object detection sub-network, 2D keypoints regression sub-network, and dimension regression sub-network. The whole system represents an end-to-end trainable network, depicted in Figure~\ref{fig:pipeline}, with sub-networks initially trained independently, switching further to joint training via introduced multi-task consistency reprojection loss function on the projected 3D keypoints and 3D bounding box corners.

\paragraph{2D object detection. }
\label{framework:2d-detection}

For 2D detection, we follow the original Mask R-CNN architecture~\cite{he2017mask}, which includes Feature Pyramid Network (FPN) \cite{lin2016feature}, Region Proposal Network (RPN) \cite{ren2015faster} and RoIAlign module \cite{he2017mask}. The RPN generates 2D anchor-boxes with a set of fixed aspect ratios and scales throughout the area of the provided feature maps, which are scored for the presence of the object of interest and adjusted. The spatial diversity of the proposed locations is processed by the RoIAlign block, converting each feature map, framed by the region of interest, into a fixed-size grid, preserving an accurate spatial location through bilinear interpolation. Followed by fully connected layers, the network splits into two feature sharing branches for the bounding box regression and object classification. During training, we utilize smooth L1 and cross-entropy loss for each task respectively, as proposed by~\cite{ren2015faster}. Though we do not directly utilize the predicted 2D bounding boxes, we have experimentally observed the 2D detection sub-network to stabilize training.

\paragraph{2D keypoint detection. }
\label{framework:2d-keypoints}

\begin{figure}[t]
 \centering
 \includegraphics[width=0.5\textwidth, height=\textheight,keepaspectratio]{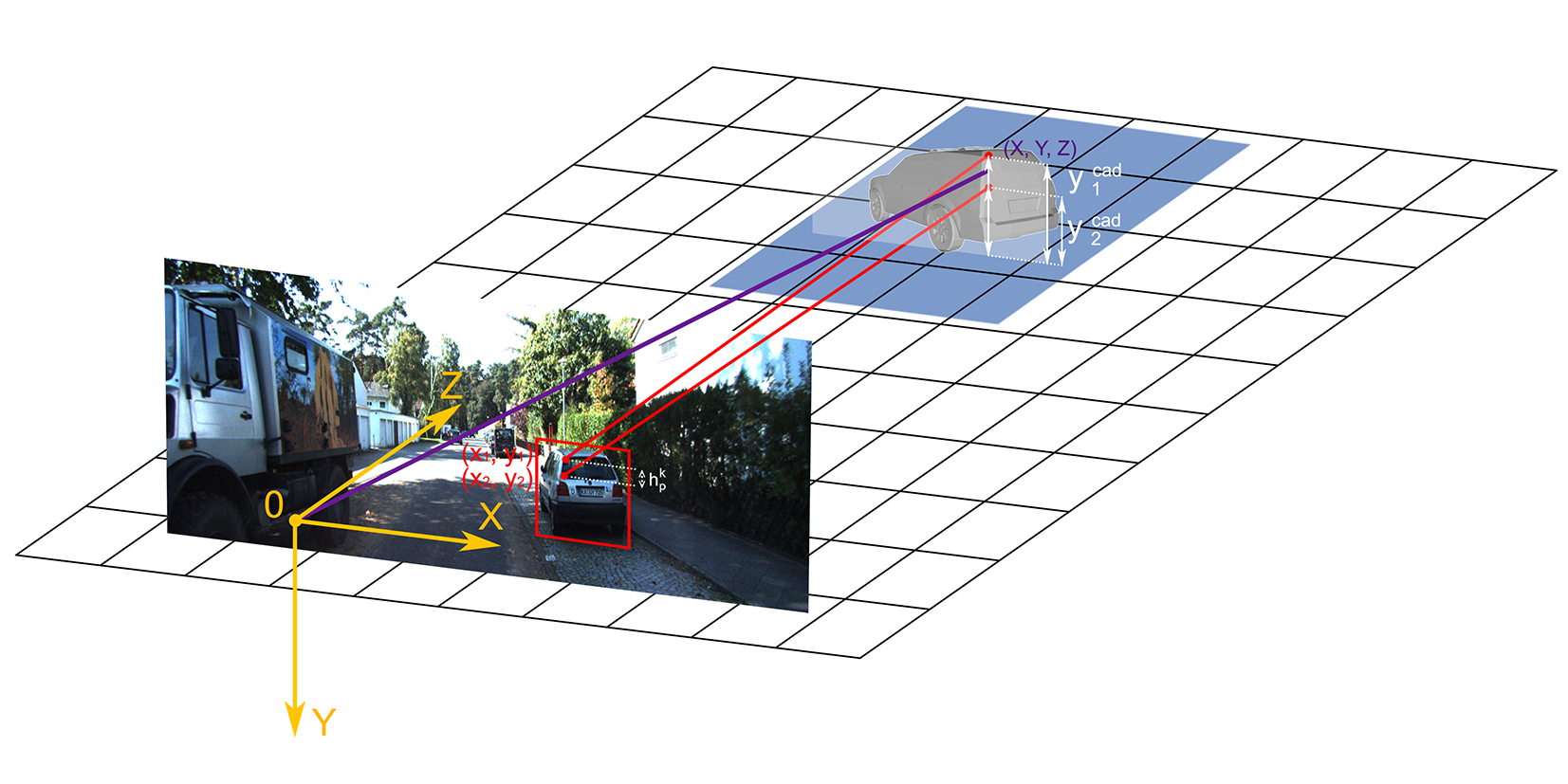}
 \caption{Geometric reasoning about instance depth. We use camera intrinsic parameters, predicted 2D keypoints and dimension predictions to "lift" the keypoints to 3D space.}
  \label{fig:keypoints}
\end{figure}

We predict coordinates and a~visibility state for each of the manually-chosen 14\,keypoints $\textbf{K} = \{(x_i, y_i)_{i=1}^{14}\}$ (\cf.~Figure~\ref{fig:keypoints} for details on our choice of 3D keypoints). 
Unlike the parameterization suggested in~\cite{he2017mask,tompson2014joint}, we directly regress on 2D coordinates of keypoints. The visibility state, determined by the occlusion and truncation of an instance, is a binary variable, and no difference between occluded, self-occluded and truncated states is made. Adding visibility estimation helps propagate information during training for visible keypoints only and acts as an auxiliary supervision for orientation sub-network.
During training, similar to our 2D object detection sub-network, we minimize the multi-task loss combining smooth L1 loss for coordinates regression and cross-entropy loss for visibility state classification, defined as:
\begin{equation}
\begin{aligned}
    &\mathcal{L}_{\text{coord}} = \sum\limits_{k=1}^K\mathbb{1}_k\cdot L_{1}^{\text{smooth}}\big(k^{\text{gt}}_{(x, y)},  k^{\text{pred}}_{(x, y)}\big)\\
    & \mathcal{L}_{\text{vis}} =  -\sum\limits_{k=1}^K \sum\limits_{j=1}^2 v^{\text{gt}}_{kj}\log v^{\text{pred}}_{kj}  \\
    &\mathcal{L}_{\text{kp}} = \mathcal{L}_{\text{coord}} + \mathcal{L}_{\text{vis}}, \\
\end{aligned}
\end{equation}
where $\mathbb{1}_k$ is the visibility indicator of $k$-th keypoint, while $k^{\text{gt}}_{(x, y)}$ and $k^{\text{pred}}_{(x, y)}$ denote ground-truth and predicted 2D coordinates, normalized and defined in a reference frame of a specific feature map after RoI-alignment.
Similarly, $v^{\text{gt}}_{kj}$ is the ground truth visibility status, while $v^{\text{pred}}_{kj}$ is the estimated probability that keypoint $k$ is visible.

\paragraph{3D dimension estimation and geometric classification. }
\label{framework:3d-dim-cad-model}

\begin{figure}[ht]
  \begin{center}
    \includegraphics[width=1.\columnwidth]{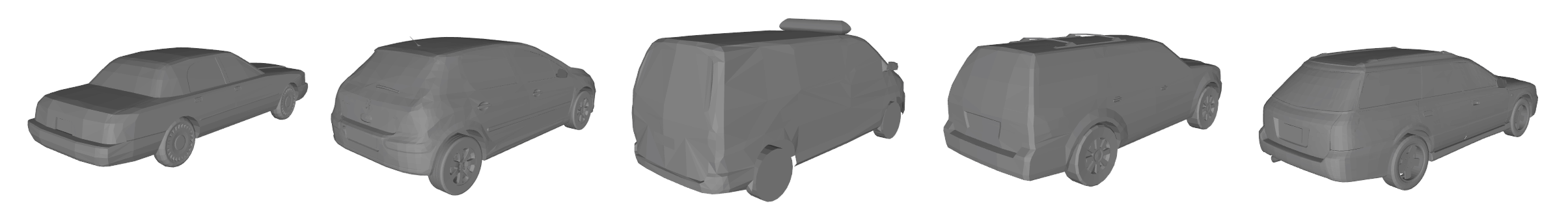}
  \end{center}
  \caption{The 5 geometric classes of instances in our work are represented by 5 3D CAD models with strongly distinct aspect ratios.}
  \label{fig:3d_cad_models}
\end{figure}

To each annotated 3D instance in the dataset, we have assigned a 3D CAD model out of a predefined set of 5~templates, obtaining 5~distinct geometric classes of instances. Used templates are presented on Figure~\ref{fig:3d_cad_models}. The assignment has been made based on the width, length and height ratios only. 
For each geometric class, we have computed mean dimensions $(\mu_w, \mu_h, \mu_l)$ over all assigned annotated 3D instances.

During training the 3D dimension estimation and geometric class selection sub-network, we utilize a multi-task loss combining cross-entropy loss (for the geometric class selection) and a smooth $L_1$ loss for dimension regression.
Instead of regressing the absolute dimensions, we predict the differences $\textbf{D}_{\text{offset}} = (\Delta w, \Delta h, \Delta l) = (w - \mu_w, h - \mu_h, l - \mu_l)$ from the mean dimensions in the log-space:
\begin{equation}
    \mathcal{L}_d((\textbf{D}^{\text{gt}}, \textbf{D}^{\text{pred}}) = L_{1}^{\text{smooth}}\big(\log(\textbf{D}^{\text{gt}}_{\text{offset}} - \textbf{D}^{\text{pred}}_{\text{offset}})\big)
\end{equation}
where $\textbf{D}^{\text{gt}}_{\text{offset}}$ and $\textbf{D}^{\text{pred}}_{\text{offset}}$ represent the ground truth and predicted offsets to the class mean values along each dimension, respectively.


\paragraph{Reasoning about instance depth. }
\label{framework:instance-depth}

We define instance depth as the depth $Z$ of a vertical plane passing through the two closest of visible keypoints, defined in the camera reference frame. To compute this depth value, we use predicted 2D keypoints, instance height (in meters), and its geometric class. First, we select two keypoints $(x_1, y_1)$ and $(x_2, y_2)$ in the image and compute their $y$-difference $h_p = |y_1 - y_2|$. We then select the corresponding two keypoints $(x^{\text{cad}}_1, y^{\text{cad}}_1)$ and $(x^{\text{cad}}_2, y^{\text{cad}}_2)$ in cad model reference frame and compute their height ratio $r_{\text{cad}} = y^{\text{cad}}_1 / y^{\text{cad}}_2$. Finally, the distance to the object $Z$ is defined from the pinhole camera model:
\begin{equation}
Z = f \cdot\dfrac{r_{\text{cad}} \cdot h}{h_p}
\end{equation}
where $f$ is a focal length of the camera, known for each frame. Figire~\ref{fig:keypoints} illustrates this computation. Depth coordinate $Z$ allows to retrieve the remaining 3D location coordinates of one of the selected keypoints, using the back-projection mapping:
\begin{equation}
   X = Z \cdot \dfrac{x_{\{1,2\}} - p_x}{f},\hspace{4mm} Y = Z \cdot \dfrac{y_{\{1,2\}} - p_y}{f}
\end{equation}
where $(p_x, p_y)$ are the camera principal point coordinates in pixels.

\paragraph{Orientation estimation. }
\label{framework:orientation}

Direct estimation of orientation $\textbf{R}$ in a camera reference frame is not feasible, as the region proposal network propagates the context within the crops solely, cutting off the relation of the crop to the image plane. 
Inspired by~\cite{MousavianCVPR2017}, we represent the global orientation as a combination of two rotations with azimuths defined as:
\begin{equation}
\theta = \theta_{\text{loc}} + \theta_{\text{ray}}
\end{equation}
where $\theta_{\text{loc}}$ is the object's local orientation within the region of interest, and $\theta_{\text{ray}}$ is a ray direction from the camera to the object center, directly found from the 3D location coordinates. We estimate $\theta_{\text{loc}}$ using a modification of the MultiBin approach~\cite{MousavianCVPR2017}. Specifically, instead of splitting the objective into angle confidence and localization parts, we discretize the angle range from \ang{0} to \ang{360} degrees into 72 non-overlapping bins and compute the probability distribution over this set of angles by a softmax layer. We train the local orientation sub-network using cross-entropy as a loss function. To obtain the final prediction for $\theta_{\text{loc}}$, we utilize the weighted mean of the bins medians ($WM({\theta}_{\text{loc}})$), adopting the softmax output as the weights. Given 3D location coordinates $(X, Z)$ of one of the keypoints and the weighted mean local orientation $WM({\theta}_{\text{loc}})$, the global orientation is defined as follows:
\begin{equation}
\theta = WM({\theta}_{\text{loc}}) +  \arctan\Big(\dfrac{X}{Z}\Big).
\end{equation}

\paragraph{3D object detection. }
\label{framework:3d-location}

To obtain the center $\textbf{C}$ of the final 3D bounding box, we use the global orientation $\mathbf{R}$ and the distance between the keypoint and the object center. For a particular CAD model, given the weight, height and length ratio between the selected keypoint and the object center $\textbf{r}_{\text{cad}} = (x^{1}_{\text{cad}} / x^{2}_{\text{cad}}, y^{1}_{\text{cad}} / y^{2}_{\text{cad}}, z^{1}_{\text{cad}} / z^{2}_{\text{cad}})$
estimated object dimensions $\textbf{D}$ and global orientation $\textbf{R}$, the location $\textbf{C}$ is predicted as
\begin{equation}
    \textbf{C} = (X, Y, Z) \pm \textbf{R} \cdot \textbf{D} \odot \textbf{r}_{\text{cad}}
\end{equation}
where $\odot$ stands for an element-wise product. Depending on the selected keypoint position (left or right, back or front, top or bottom side of the object), a sign is chosen for each dimension.

\begin{figure}[t]
 \centering
 \includegraphics[width=0.5\textwidth, height=\textheight,keepaspectratio]{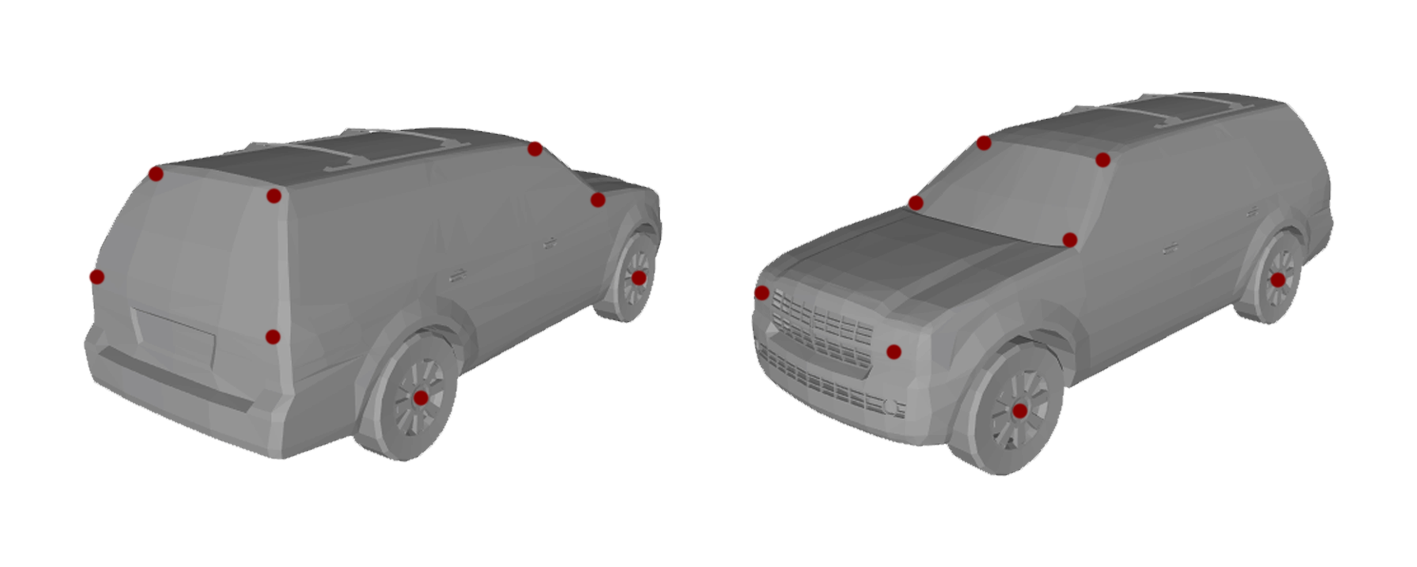}
 \label{fig:annotation}
 \caption{An example of our annotation of 3D keypoints. We label 4 centers of wheels, 8 corners of front and back windshields per each, and 2 centers of headlights. Per each 3D CAD model, we form the annotation so as to ensure that the two keypoints on each of the windshields form near-vertical lines. This is required to ensure accuracy of our geometric reasoning framework.}
\end{figure}

\paragraph{Multi-head reprojection consistency loss. }
\label{framework:reprojection-loss}

Except for shared convolutional backbone, each sub-network is independent of its neighbors and unaware of other predictions, though the geometric components are strongly interrelated. To provide consistency between the network branches, we introduce a loss function which integrates all the predictions. 3D coordinates in a CAD model coordinate system of the keypoints from the set $\textbf{K}$ are scaled using $\textbf{D}$, rotated using $\textbf{R}$, translated using $\textbf{C}$, and back-projected into the image plane via camera projection matrix to obtain 2D keypoint coordinates and compare with the ground truth values. A similar approach is applied to the eight corners of the 3D bounding box obtained from 3D detection and orientation estimates, to ensure that they fit tightly into the ground truth 2D bounding box after back-projection. In all cases, we use the smooth L1 loss during training.



\section{Experiments}
\label{sec:exper}

\subsection{Experimental setup}
\label{exper:setup}

 \begin{table*}[ht]
\centering
\resizebox{0.60\textwidth}{!}{
\begin{tabular}{|c|c|c|c|c|c|c|}
\hline
\multirow{0}{*}{
Method} & \multicolumn{3}{c|}{IoU = 0.5} & \multicolumn{3}{c|}{IoU = 0.7} \\ \cline{2-7} 
 & Easy & Moderate & Hard & Easy & Moderate & Hard \\ \hline
Mono3D & 25.19 & 18.20 & 15.52 & 2.53 & 2.31 & 2.31 \\ \hline
OFT-Net & - & - & - & 4.07 & 3.27 & 3.29 \\ \hline
MonoGRNet & 50.51 & \textbf{36.97} & \textbf{30.82} & 13.88 & \textbf{10.19} & \textbf{7.62} \\ \hline
MF3D & 47.88 & 29.48 & \underline{26.44} & 10.53 & 5.69 & \underline{5.39} \\ \hline
Ours & 48.81 & 30.17 & 20.07 & 11.91 & 6.64 & 4.28 \\ \hline
Ours (+loss) & \textbf{50.82} & \underline{31.28} & 20.21 & \textbf{13.96} & \underline{7.37} & 4.54 \\ \hline
\end{tabular}}
\caption{\textbf{3D detection performance:} Average Precision of 3D bounding boxes on KITTI val set. The best score is in bold, the second best underlined. Ours (+loss) indicates a base network setup, trained with a consistency reprojection loss.}
\label{table:results}
\end{table*}

\paragraph{Dataset. }
We train and evaluate our approach using the KITTI 3D object detection benchmark dataset. For the sake of comparison with state-of-the-art methods, we follow the setup presented in~\cite{Chen2016CVPR},  which provides 3712 and 3769 images for training and validation, respectively, along with the camera calibration data. To extend KITTI dataset with assignment of geometric classes using CAD models and keypoints 2D coordinates, we employ the approach and data provided in \cite{xiang_cvpr15}. Depending on the ratios between height, length and width, each car instance is assigned with one out of 5 CAD model classes from a predefined set of CAD templates, presented on Figure~\ref{fig:pipeline}. We manually annotated each CAD model with the keypoint locations. Figure~\ref{fig:keypoints} displays an example of the annotated keypoints, most of which are a common choice~\cite{Everingham10thepascal} due to their interpretability, such as the car's edges, carcass, \etc.; we also included corners of windshields to deal with the height of each instance. To obtain 2D coordinates of the keypoints, we back-projected CAD models from 3D space to the image plane using ground truth location, dimension and rotation values. Simultaneous projection of all 3D CAD models on a scene provides us with a depth ordering mask, allowing for defining the visibility state of each keypoint.

\paragraph{Network architecture. }
 We utilize Mask R-CNN with a Feature Pyramid Network \cite{lin2016feature}, based on a ResNet-101 \cite{he2015deep} as our backbone network for the multi-level feature extraction. Instead of the higher resolution $14 \times 14$ and $28 \times 28$ feature maps in the original architecture, we stack the same amount of $3 \times 3$ kernels, followed by a fully connected layer to predict 2D normalized coordinates and visibility states for each of the 14 keypoints. From the same feature maps, we branch a fully-connected layer predicting local orientation in bins of \ang{5} each, totaling 72 output units. The feature sharing between keypoints and local orientation was found crucial for network performance, as both characteristics imply similar geometric reasoning. In parallel to 2D detection and 2D keypoints estimation, we create a sub-network of a similar architecture for dimension regression and classification into geometric classes. The remaining components, including RPN, RoIAlign, bounding box regression and classification heads, are implemented following the original Mask R-CNN design. For instance depth retrieval we use only four pairs of keypoints: corners of the front and rear windows. Other keypoints are used for additional supervision in consistency loss calculation during training. 
 
\paragraph{Training our model. }

We set hyperparameters following Mask R-CNN work \cite{he2017mask}. The RPN anchor set covers five scales, adjusting them to the values of 4, 8, 16, 32, 64, and three default aspect ratios. Each mini-batch consists of 2 images, producing 256 regions of interest, with a positive to negative samples ratio set 1:3, to achieve class sampling balance during training. Any geometric augmentations over the images are omitted, solely applying image padding to meet the network architecture requirements. ResNet-101 is initialized with the weights pre-trained on Imagenet~\cite{imagenet_cvpr09}, and frozen during further training steps. We first train the 2D detection and classification sub-network for 100K iterations, adopting Adam optimizer with a learning rate of $10^{-4}$ throughout the training, setting weight decay of 0.001 and momentum of 0.9. Then 2D keypoints and local orientation are trained for 50K iterations. Finally, enabling the multi-head consistency loss, the whole network is trained in an end-to-end fashion for 50K iterations. We combine losses from all of the network outputs, weighting them equally.

\paragraph{Evaluation metrics. }
We evaluate the network under the conventional KITTI benchmark protocol, which enables comparison across approaches. Car category is the sole subject of our focus. By default, KITTI settings require evaluation in 3 regimes: easy, moderate and hard, depending on the instance difficulty of a potential detection. 3D bounding box detection performance implies 3D Average Precision (AP3D)  evaluation, setting Intersection over Union (IoU) threshold to 0.5 and 0.7.

\subsection{Experimental results}
\label{exper:results}

\begin{figure*}[ht!]
  \centering
  \subcaptionbox*{(a)}[.33\linewidth][c]{%
    \includegraphics[width=.33\linewidth]{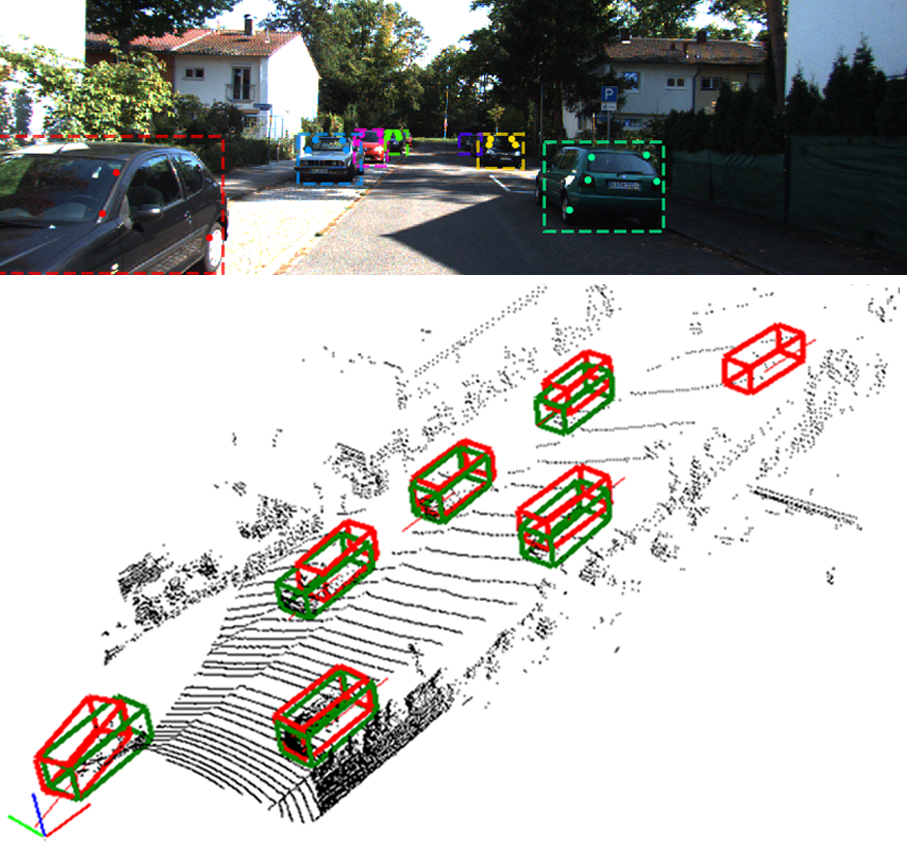}}
  \subcaptionbox*{(b)}[.33\linewidth][c]{%
    \includegraphics[width=.33\linewidth]{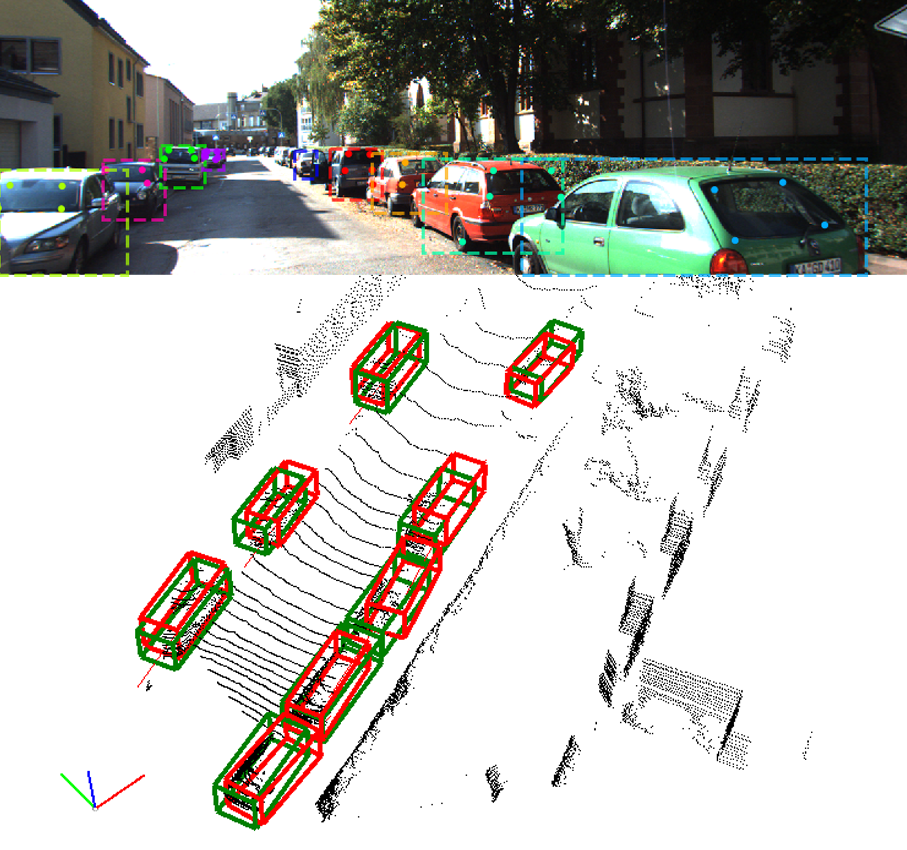}}
  \subcaptionbox*{(c)}[.33\linewidth][c]{%
    \includegraphics[width=.33\linewidth]{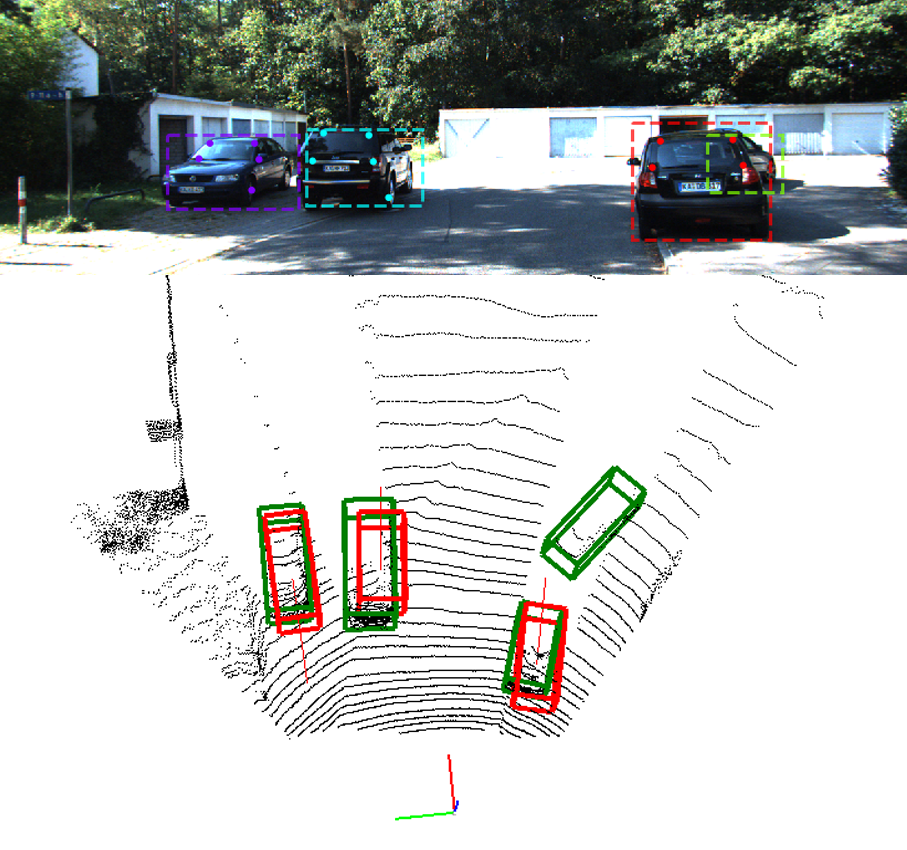}}

  \bigskip

  \subcaptionbox*{(d)}[.33\linewidth][c]{%
    \includegraphics[width=.33\linewidth]{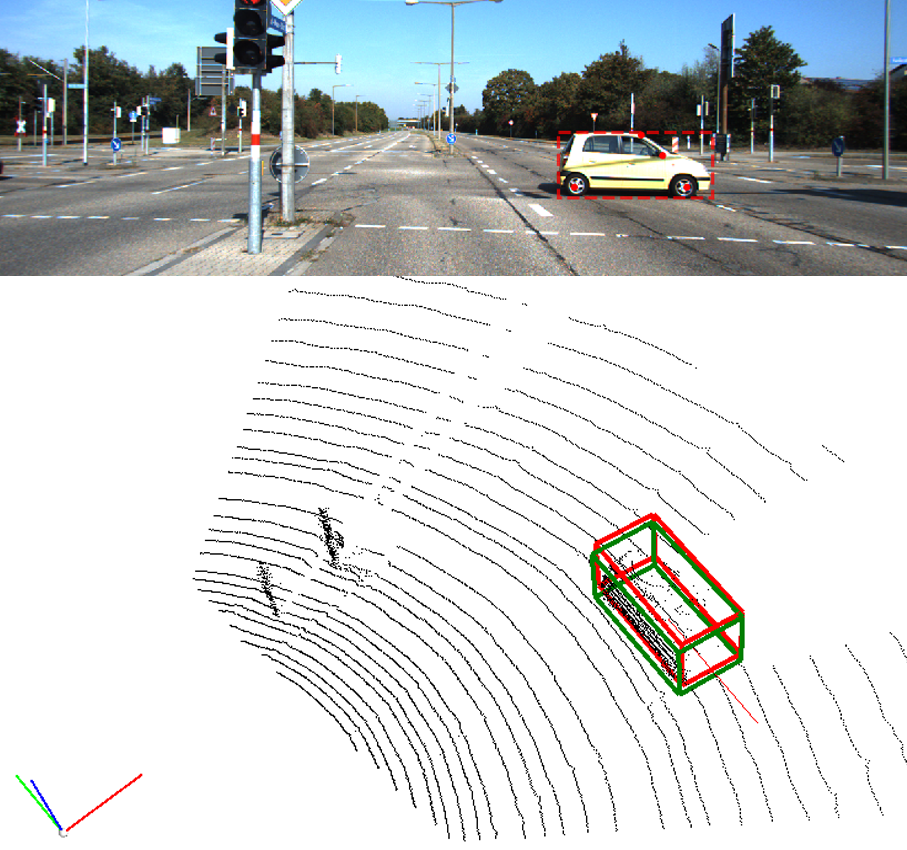}}
  \subcaptionbox*{(e)}[.33\linewidth][c]{%
    \includegraphics[width=.33\linewidth]{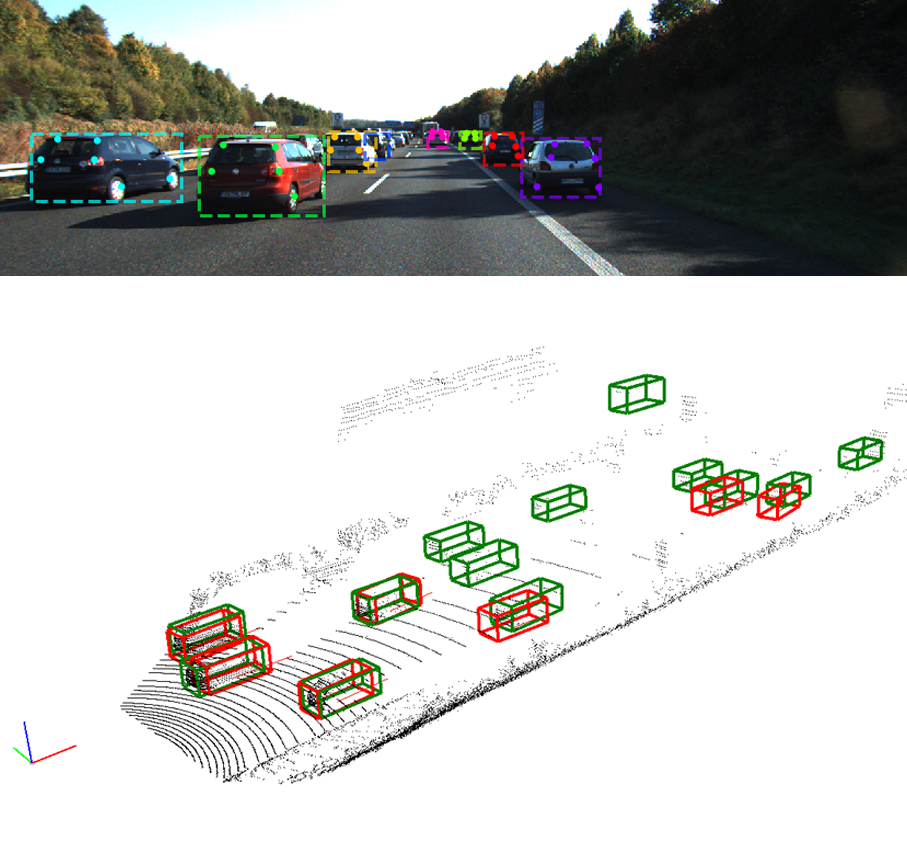}}
  \subcaptionbox*{(f}[.33\linewidth][c]{%
    \includegraphics[width=.33\linewidth]{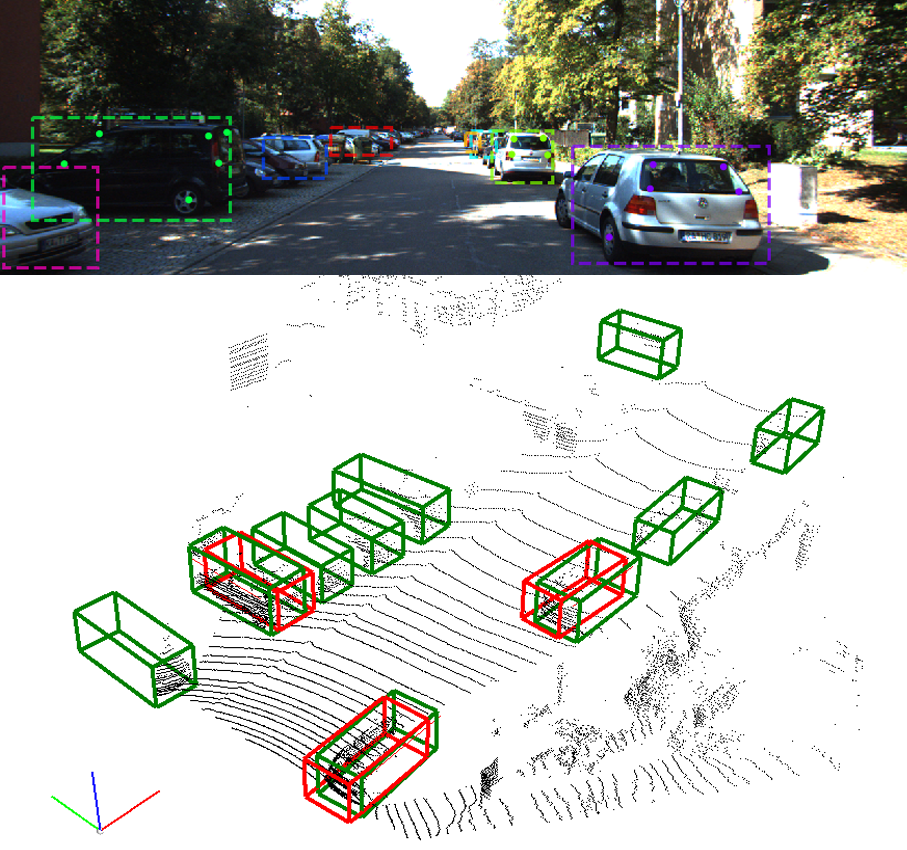}}
  \caption{Qualitative results. The upper part of each sub-figure contains 2D detection inference, including 2D bounding and 2D locations of the visible keypoints. Each instance and its keypoints are displayed their distinctive color. The lower part visualizes the 3D point cloud, showing the camera location as the colored XYZ axes. Green and red colors stand for the ground truth and predicted 3D bounding boxes respectively. The scenes were selected to express diversity in complexity and cars positioning \wrt. the camera.}
  \label{fig:inference}
\end{figure*}

\paragraph{3D object detection. }
We compare the performance with 4 monocular 3D object detection methods: Mono3D \cite{Chen2016CVPR}, OFT-Net \cite{roddick2018orthographic}, MonoGRNet \cite{qin2018monogrnet} and MF3D \cite{xu2018multi}, which reported their results on the same validation set for the car class. We borrow the presented average precision numbers from their published results. The results are reported in Table ~\ref{table:results}. The experiments show that our approach outperforms state-of-the-art methods on the easy subset by a small margin while remaining the second best on the moderate subset. This observation aligns with our intuition that visible salient features such as keypoints are crucial to the success of 3D pose estimation. For the moderate and hard images, 2D keypoints are challenging to robustly detect due to the high occlusion level or the low resolution of the instance. We also measure the effect of the reprojection consistency loss on our network performance, observing a positive effect of our loss function. 

\paragraph{3D bounding box and global orientation estimation. }
We follow the experiment presented in \cite{qin2018monogrnet}, evaluating the quality of the 3D bounding boxes sizes estimation, as well as the orientation in a camera coordinate system. The mean errors of our approach, along with \cite{qin2018monogrnet, Chen2016CVPR}, borrowed from their work, are presented in Table \ref{table:dim_orient}.

\begin{table}[h!]
\centering
\resizebox{0.48\textwidth}{!}{
\begin{tabular}{|c|c|c|c|c|}
\hline
\multirow{0}{*}{Method} & \multicolumn{3}{c|}{Size (m)} & \multirow{0}{*}{Orientation (rad)} \\ \cline{2-4}
 & Height & Width & Length &  \\ \hline
Mono3D & 0.172 & 0.103 & 0.504 & 0.558 \\ \hline
MonoGRNet & \textbf{0.084} & \textbf{0.084} & {\ul 0.412} & 0.251 \\ \hline
Ours & 0.115 & 0.107 & 0.516 & {\ul 0.215} \\ \hline
Ours (+loss) & {\ul 0.101} & {\ul 0.091} & \textbf{0.403} & \textbf{0.191} \\ \hline
\end{tabular}}
\caption{\textbf{3D bounding box and orientation mean errors:} The best score is in bold, the second best underlined.}
\label{table:dim_orient}
\end{table}
\noindent Though, the sizes of the 3D bounding boxes do not differ severely among the approaches, due to the estimating the offset from the median bounding box, the orientation estimation results differ significantly. Since we retrieve global orientation via geometric reasoning, learning local orientation from 2D image features, the network provides more accurate predictions, in contrast to obtaining orientation from the regressed 3D bounding box corners.

\paragraph{Qualitative results. }
We provide a qualitative illustration of the network performance in Figure~\ref{fig:inference}, displaying six road scenes with distinct levels of difficulty. In typical cases, our approach produces accurate 3D bounding boxes for all instances, along with the global orientation and  3D location. Remarkably, the truncated objects can also be successfully detected, given that only one pair of keypoints hits the image. Some hard cases, i.e. (e) and (f), primarily consist of objects that are distant, highly occluded or even invisible on the image. We believe such failure cases to be a common limitation of monocular image processing methods.

\section{Conclusions}
\label{sec:conclusion}

In this work, we presented a novel deep learning-based framework for monocular 3D object detection combining well-known detectors with geometric reasoning on keypoints. We proposed to estimate correspondences between the detected 2D keypoints and their 3D counterparts annotated on the surface of 3D CAD models to solve the object localization problem. Results of the experimental evaluation of our approach on the subsets of the KITTI 3D object detection benchmark demonstrate that it outperforms the competing state-of-the-art approaches when the target objects are clearly visible, leading us to hypothesize that dense depth estimation is redundant for 3D detection in some instances. We have demonstrated our multi-task reprojection consistency loss to significantly improve performance, in particular, the orientation of detections.

\section*{Acknowledgement}
\noindent E. Burnaev and A. Artemov were supported by the Russian Science Foundation under Grant 19-41-04109.

{\small
\bibliographystyle{ieee}
\bibliography{egbib}
}

\end{document}